\documentclass{article}

\usepackage{arxiv}
\usepackage{amsmath,amssymb,amsfonts}
\usepackage{algorithmic}
\usepackage{graphicx}
\usepackage{textcomp}
\usepackage{xcolor}
\def\BibTeX{{\rm B\kern-.05em{\sc i\kern-.025em b}\kern-.08em
    T\kern-.1667em\lower.7ex\hbox{E}\kern-.125emX}}

\usepackage[bookmarks=false, pdfencoding=auto, pdfborder={0 0 0}, colorlinks=false]{hyperref}

\usepackage[colorinlistoftodos,prependcaption,textsize=tiny]{todonotes}

\usepackage[inkscapelatex=false]{svg}
\usepackage{pdfpages}

\usepackage[acronym]{glossaries}
\newacronym{GUI}{GUI}{Graphical User Interface}
\newacronym{MVC}{MVC}{Model-View-Controller}
\newacronym{FSM}{FSM}{Finite State Machine}
\newacronym{ML}{ML}{Machine Learning}
\newacronym{QoS}{QoS}{Quality of Service}

\newacronym{FCU}{FCU}{Flight Control Unit}
\newacronym{RC}{RC}{Remote Control}

\newacronym{SDR}{SDR}{Software Defined Radio}
\newacronym{GNSS}{GNSS}{Global Navigation Satellite System}

\newacronym{EO}{EO}{Electro-Optical}
\newacronym{IR}{IR}{Infrared}
\newacronym{RF}{RF}{Radio Frequency}
\newacronym{IMU}{IMU}{Inertial Measurement Unit}

\newacronym{EOL}{EOL}{End of Life}
\newacronym{ROS1}{ROS 1}{Robot Operating System 1}
\newacronym{ROS2}{ROS 2}{Robot Operating System 2}
\newacronym{DDS}{DDS}{Data Distribution Service}

\newacronym{UGV}{UGV}{Unmanned Ground Vehicle}
\newacronym{UAV}{UAV}{Unmanned Aerial Vehicle}
\newacronym{UAS}{UAS}{Unmanned Aerial System}
\newacronym{cUAS}{C-UAS}{Counter Unmanned Aerial System}
\newacronym{C2}{C2}{Command and Control}
\newacronym{SPOF}{SPOF}{Single Point of Failure}
\newacronym{WLAN}{WLAN}{Wireless Local Area Network}
\newacronym{5G}{5G}{Fifth Generation Mobile Network}

\usepackage[backend=biber, style=ieee, sorting=none]{biblatex} 
\title{A Modular and Scalable System Architecture for Heterogeneous UAV Swarms Using ROS 2 and PX4-Autopilot}
\renewcommand{\shorttitle}{}

\hypersetup{
pdftitle={A Modular and Scalable System Architecture for Heterogeneous UAV Swarms Using ROS 2 and PX4-Autopilot},
pdfsubject={UAS, UAV Swarm, ROS 2, PX4-Autopilot, Docker},
pdfauthor={Robert Pommeranz},
pdfkeywords={UAS, UAV Swarm, ROS 2, PX4-Autopilot, Docker},
}

\author{
 Robert Pommeranz \\
  Electrical Measurement Engineering\\
  Helmut-Schmidt-University\\
  Hamburg, Germany \\
  \texttt{robert.pommeranz@hsu-hh.de} \\
   \And
 Kevin Tebbe \\
  Electrical Measurement Engineering\\
  Helmut-Schmidt-University\\
  Hamburg, Germany \\
  \texttt{kevin.tebbe@hsu-hh.de} \\
  \And
 Ralf Heynicke \\
  Electrical Measurement Engineering\\
  Helmut-Schmidt-University\\
  Hamburg, Germany \\
  \texttt{ralf.heynicke@hsu-hh.de} \\
  \And
 Gerd Scholl \\
  Electrical Measurement Engineering\\
  Helmut-Schmidt-University\\
  Hamburg, Germany \\
  \texttt{gerd.scholl@hsu-hh.de} \\
}

\begin{document}
\maketitle

\begin{abstract}
    \footnote{This is the author's version of a paper that has been accepted for presentation at the 12th IEEE International Conference on Mechatronics and Robotics Engineering (ICMRE), to be held in Oldenburg, Germany, on March 2–4, 2026.}In this paper a modular and scalable architecture for heterogeneous swarm-based \glspl{cUAS} built on PX4-Autopilot and \gls{ROS2} framework is presented.
    The proposed architecture emphasizes seamless integration of hardware components by introducing independent \gls{ROS2} nodes for each component of a \gls{UAV}. Communication between swarm participants is abstracted in software, allowing the use of various technologies without architectural changes.
    Key functionalities are supported, e.g. leader following and formation flight to maneuver the swarm. The system also allows computer vision algorithms to be integrated for the detection and tracking of \glspl{UAV}. Additionally, a ground station control is integrated for the coordination of swarm operations.
    Swarm-based \gls{UAS} architecture is verified within a Gazebo simulation environment but also in real-world demonstrations.
\end{abstract}


\glsresetall

\section{Introduction}
Swarm-based \glspl{UAS} pose significant security challenges to critical infrastructure, including government facilities, airports, and military installations. Unlike individual drones, swarms can coordinate to evade conventional defenses, execute complex missions, and collect sensitive information, thereby amplifying the potential impact of an attack.
Existing \gls{cUAS} solutions are primarily designed for single drone threats and often lack the capability to address the speed, coordination, and adaptability of swarms~\cite{tebbeEngineeringPerspectiveSmall2025,chipadeAerialSwarmDefense2023}.
To counter these evolving threats, \gls{cUAS} platforms must enable coordinated operation across multiple sensors and actuators.
Intelligent algorithms are essential for detecting and tracking both individual \glspl{UAV} and collective swarms in real-time under dynamic conditions, in order to devise effective defense strategies~\cite{brambillaSwarmRoboticsReview2013}. Comprehensive surveys such as~\cite{kangProtectYourSky2020} provide an overview of \glspl{cUAS}, detailing key technologies, architectures, and market trends, and emphasize the importance of holistic approaches that integrate sensing, \gls{C2}, and mitigation systems. In~\cite{shiAntiDroneSystemMultiple2018}, the authors present a multi-technology anti-drone system architecture and discuss its implementation and associated challenges. They demonstrate the effectiveness of combining passive surveillance technologies for detection, localization, and \gls{RF} jamming.
\\
Employing a multi-sensor strategy integrating data from sources such as \gls{EO} / \gls{IR} cameras, radar, and \gls{RF} analyzers alongside \gls{ML} algorithms—improves detection accuracy, reduces false alarms, and enhances the identification of swarm behaviors in diverse environments. Fusing multiple sensor modalities also increases system resilience and adaptability, enabling early threat detection, swarm trajectory forecasting, and more effective countermeasures. This approach underscores the importance of modularity, given the diverse functions and algorithms that must operate concurrently within robotic systems~\cite{jahnConceptsModularSystem2019}.
\\
Here, a modular hardware and software architecture for \gls{cUAS} applications, utilizing \gls{ROS2} to enable decentralized, real-time swarm operations, will be presented. 
Each \gls{UAV} operates containerized \gls{ROS2} nodes for hardware abstraction, sensor integration, and mission management, facilitating reproducible and scalable deployments. The architecture separates flight control, sensor/actuator signal processing, and swarm control into independent modules, enhancing exchangeability, maintainability and adaptability to different hardware platforms. Also, advanced swarm behaviors including leader-follower roles with formations are supported to maneuver the swarm.
Algorithms for processing sensor signals including the camera are implemented as separate \gls{ROS2} nodes. Communication relies on \gls{DDS} for reliable, low-latency data exchange, while a ground station provides centralized monitoring and control.
\\
This paper is structured as follows: Section \ref{sec:related_work} provides related work. Section \ref{sec:system_architecture} outlines the proposed architecture, while
Section \ref{sec:implementation} describes its implementation in detail. Finally, Section \ref{sec:conclusion} concludes the paper and outlines future work.

\section{Related Work}
\label{sec:related_work}
The usage of \gls{ROS2} in \glspl{UAS} has been a significant area of research, enabling decentralized, scalable, and reusable robotic architectures. Early implementations predominantly utilized \gls{ROS1}, but recent studies have increasingly shifted toward \gls{ROS2} due to the \gls{EOL} of \gls{ROS1} on May 31st, 2025. Key advancements in \gls{ROS2} include an enhanced middleware architecture based on \gls{DDS} and a decentralized paradigm that eliminates the need for a master node~\cite{mazzeoTROSProtectingHumanoids2020,maruyamaExploringPerformanceROS22016}.
\gls{DDS} is a publish-subscribe middleware that provides a data-centric communication model, enabling efficient and reliable data exchange between distributed systems. Its definition~\cite{OMGOmgData2015} is an open standard by the Object Management Group (OMG) and is widely used in various industries, including aerospace, automotive, and healthcare. DDS provides features such as data distribution, \gls{QoS} settings, and real-time communication.
\\
The MRS Drone platform introduces a modular \gls{UAV} system built on \gls{ROS1}, emphasizing hardware modularity and multi-robot deployment~\cite{hertMRSDroneModular2023}. The authors in~\cite{kaiserROS2SWARMROS22022} present ROS2SWARM as a reusable ROS2-based library for swarm robot behaviors, supporting multiple \glspl{UGV} and demonstrating the versatility of \gls{ROS2} for swarm robotics. 
\\
Aerostack2 is an open-source ROS2-based framework for multi-robot aerial systems, introducing a modular plugin architecture and platform independence~\cite{fernandez-cortizasAerostack2SoftwareFramework2024}. 
\\
CrazyChoir provides a lightweight, modular swarm control framework using \gls{ROS2} for Crazyflie nano-UAVs~\cite{pichierriCrazyChoirFlyingSwarms2023a}. CrazySwarm2\footnote{\url{https://imrclab.github.io/crazyswarm2}} is the ROS2-based successor to CrazySwarm, designed for large-scale, coordinated control of Crazyflie nano-quadcopters, leveraging ROS2's decentralized architecture and DDS-based middleware~\cite{preissCrazyswarmLargeNanoquadcopter2017}.
\\
However, to the best of our knowledge, a containerized solution in combination of \gls{ROS2}, PX4-Autopilot and heterogeneous swarm vehicles in a \gls{cUAS} context has not been previously addressed in the literature. The presented architecture integrates these technologies to provide a modular platform for further research.

\section{System Architecture}
\label{sec:system_architecture}
\begin{figure}[tb]
	\centering
	\includegraphics[width=\linewidth]{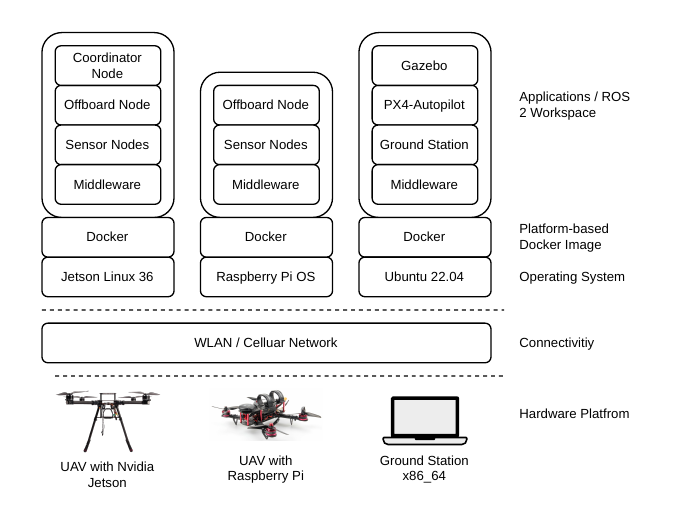}
	\caption{Software architecture of a heterogeneous \gls{UAS} integrating multiple companion computers and modular software components. The ground station enables centralized swarm control and real-time state visualization, when autonomous operations are not feasible.}
	\label{fig:system_architecture_single_uav}
\end{figure}

This section presents the architecture of the proposed swarm-based \gls{cUAS}, using PX4-Autopilot based \glspl{UAV} in combination with the \gls{ROS2} framework. The current hardware platform comprises three distinct \gls{UAV} classes, GenericUAVModel, ObservationUAVModel and CoordinatorUAVModel, selected based on payload capacity, maneuverability, and required onboard computational resources and should be open for future expansion. The software stack is structured as a collection of loosely coupled, reusable components, each implemented as an independent \gls{ROS2} node. This design enables flexible configuration of each \gls{UAV} with various sensors and actuators to meet specific mission requirements.
Fig.~\ref{fig:system_architecture_single_uav} illustrates the software architecture for the \gls{UAS}, highlighting the integration of various companion computers, such as the Nvidia Jetson Orin NX, Raspberry Pi  (AArch64) and the x86\_64 as ground station.
On the left side of the figure a \gls{UAV} equipped with a companion computer running Jetson Linux 36 is shown. Within the Docker container, the system hosts the \gls{DDS} middleware and a \gls{ROS2} workspace, which includes the PX4-Autopilot and relevant sensor nodes, depending on the specific configuration. The Offboard Node manages communication between the \gls{UAV} and the PX4-Autopilot flight controller, enabling Offboard mode for external control via commands from the companion computer. The coordinator node is responsible for managing the swarm's leader-follower roles and formation flights. In the center column of the figure, a \gls{UAV}, equipped with a Raspberry Pi 4 CM is shown, which can be used for less resource intensive algorithms.
The right side of the figure illustrates the ground station, which can be implemented on a laptop or desktop computer. The ground station runs a \gls{GUI} to control the swarm and visualize the current state of sensors and the vehicle.
Derived classes extending GenericUAVModel implement specialized features for different \gls{UAV} types, such as the \mbox{ObservationUAVModel} for surveillance missions.
The main classes and their functionalities are described below:
\paragraph{GenericUAVModel} This class initializes the \gls{ROS2} node and manages the connection to the PX4-Autopilot flight controller. In addition, this class holds a \gls{FSM} for swarm operations or individual missions and stores general swarm parameters.

\paragraph{ObservationUAVModel} Extending the GenericUAVModel class, this model incorporates dedicated publish and subscribe methods to interface seamlessly with three specific \gls{ROS2} nodes. These nodes handle gimbal control, camera management, and video streaming. Onboard processing of video streams is performed by the companion computer, which then publishes the processed streams to the ground station.

\paragraph{CoordinatorUAVModel} The CoordinatorUAVModel is responsible for managing the swarm's leader-follower roles and formation flights. It can manage the swarm's trajectory and ensure that all participants adhere to the defined formation. The CoordinatorUAVModel may be implemented as a ground station on a laptop providing an optional \gls{GUI}, or integrated onto a \gls{UAV} for decentralized swarm management, allowing any \gls{UAV} to assume the coordinator role if needed.

In summary, the proposed architecture addresses the challenges outlined in Section~\ref{sec:related_work} by emphasizing modularity, scalability, and flexibility. The use of loosely coupled software and \gls{ROS2} nodes enables the integration of diverse hardware and software components, supporting heterogeneous swarm based \gls{UAS} depending on the mission requirements.

\section{Implementation}
\label{sec:implementation}
This section outlines the implementation of the architecture, including the hardware platform, swarm communication, containerization, and the \gls{ROS2} software stack.
\subsection{Hardware Platform}
\begin{figure}
	\centering
	\includegraphics[width=\linewidth]{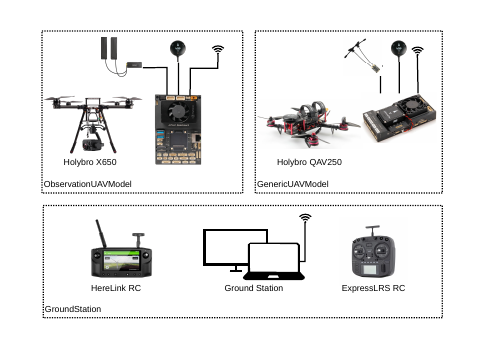}
	\caption{Potential \gls{UAS} configuration: The ObservationUAV with a multi-sensor camera. A GenericUAVModel with a Raspberry Pi CM 4. The ground station is a laptop with a \gls{GUI} to control the swarm. As a fallback solution, each \gls{UAV} can be manually controlled over a \gls{RC}, which is directly connected to the \gls{FCU}.}
	\label{fig:implementation_hardware_platform}
\end{figure}

In Fig.~\ref{fig:implementation_hardware_platform}, the platform is shown, which consists of two different hardware types depending on the type companion computer and payload requirements. The ObservationUAVModel is implemented on a Holybro X650 frame with Pixhawk Jetson Baseboard stacked with Pixhawk 6X and Jetson Orin NX 16\,GB. A multi-sensor camera with a gimbal or other cameras are attached to the companion computer. A lite version, referred to as the GenericUAVModel, is realized by a Raspberry Pi 4 as a computation module, which can be used for smaller \glspl{UAV}.
For safety reasons there is also always the possibility to control each \gls{UAV} manually over \gls{RC}. A Herelink Airunit or ExpressLRS is used, which is directly connected to the \gls{FCU}. The \gls{GNSS} is used for global position information in PX4-Autopilot Offboard Mode.

\subsection{Swarm Communication}
The swarm primarily communicates over a private \gls{WLAN} network, where all devices have access to the \gls{ROS2} topics. Such a setup allows for low-latency communication between the \glspl{UAV} and the ground station. Using \gls{WLAN} as the physical layer for communication has the advantage that one can transition from a simulation environment to a real-world environment very quickly and with minimal effort, as \gls{WLAN} modules can be easily integrated into the hardware or are already built into the controllers. In the future, this layer can be replaced with another technology, such as a cellular network (e.g., \gls{5G}), to provide more robust and reliable communication in environments with high interference. The choice of technology depends on the specific requirements of the application and the available infrastructure.

\subsection{Containerization}
To ensure that the docker container can be used on different hardware platforms, here Nvidia Jetson, Raspberry Pi or any x86\_64 based system, a slightly adjusted container with the appropriate docker tag is built with the necessary dependencies and configurations for each platform. The process of containerization has the advantage of simplifying and accelerating deployment, while ensuring consistent operation of the software stack across all platforms.
This approach enables the same \gls{ROS2} packages to be used across different hardware configurations, ensuring compatibility and reducing the risk of software conflicts.
As shown in Fig.~\ref{fig:system_architecture_single_uav}, the container includes all necessary libraries and dependencies for the \gls{ROS2} software stack, including the PX4-Autopilot, camera drivers, and other required packages. 

\subsection{\gls{ROS2} Software Stack}
The software stack is developed using Python 3.10 and \gls{ROS2} Humble, encapsulated within a Docker container for portability and consistency. The primary \gls{ROS2} packages utilized in this implementation include:
\begin{itemize}
	\item \textbf{px4\_offboard}: Provides the core functionalities for controlling the UAV, including flight control, state management, and communication with the PX4-Autopilot.
	\item \textbf{camera\_gimbal\_node}: \glspl{UAV} equipped with camera gimbals include this node in the \gls{ROS2} launch file. It enables remote gimbal control, lens control, and receiving video streams via ROS topics.
	\item \textbf{camera\_stream\_node}: Handles the streaming of camera data from the UAV to the ground station. This package processes video streams onboard the companion computer and publishes the processed streams to the ground station.
\end{itemize}

\subsection{Ground Station}
\begin{figure}[tb]
	\includegraphics[width=\linewidth]{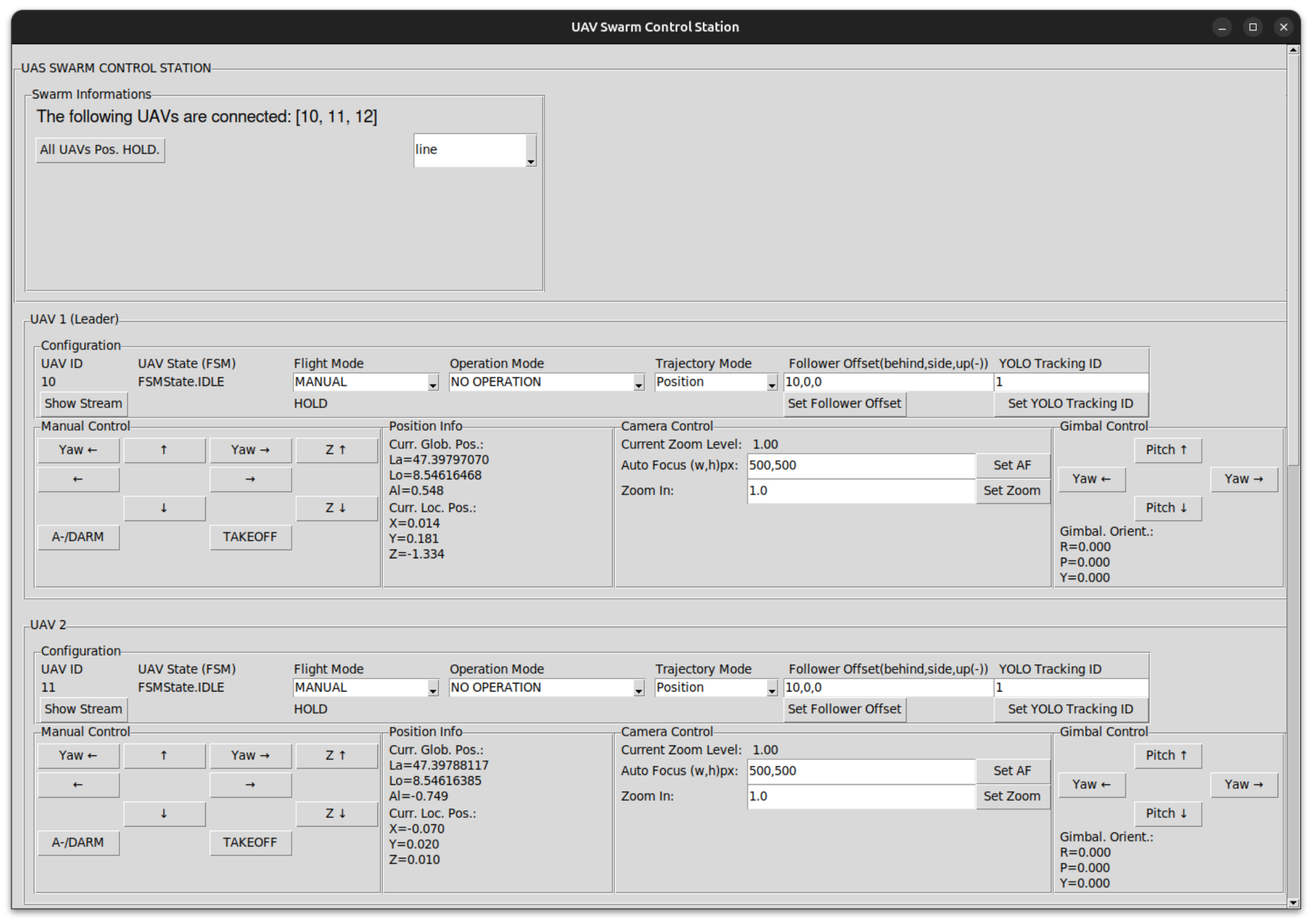}
	\centering
	\caption{Ground station with \gls{GUI} to control and display the current state of the swarm.}
	\label{fig:groundstation_gui}
\end{figure}

Fig.~\ref{fig:groundstation_gui} illustrates the \gls{GUI} of the ground station, which enables intuitive interaction with the main functions of the CoordinatorModel. The coordinator automatically detects and manages new swarm members, updating the interface in real-time. The upper section of the \gls{GUI} allows users to select and configure swarm formations, while the lower section displays each connected \gls{UAV}, providing individual control options. The software architecture is based on the Model-View-Controller (\gls{MVC}) pattern, ensuring a clear separation between the user interface and application logic. This modular design simplifies future migration to alternative GUI frameworks beyond Tkinter.

\section{Conclusion}
\label{sec:conclusion}
In this paper, we presented a software architecture for \gls{cUAS} applications, leveraging the capabilities of \gls{ROS2} and PX4-Autopilot. By deploying docker containers on each swarm participant, we ensured consistent and reproducible deployments across diverse hardware platforms, including Nvidia Jetson, Raspberry Pi and x86\_64 based systems for simulation. The architecture supports seamless communication inside the network and can utilize \gls{WLAN}, cellular networks or other wireless technologies.
A modular design facilitates the integration of diverse \gls{UAV} models and payloads, thereby enabling the system to accommodate various mission profiles and operational requirements.
In conclusion, the proposed architecture establishes a robust foundation for developing advanced swarm-based \gls{cUAS} applications and can adapt to the evolving challenges in unmanned systems. Future work will focus on enhancing the system's capabilities, including transitioning towards a private 5G network for improved communication reliability and robustness in congested environments where traditional wireless networks may be insufficient. This approach ensures long-term adaptability and scalability, allowing the system to remain effective in rapidly changing operational and technological landscapes.

\printbibliography

%
%
\end{document}